\documentclass[journal]{IEEEtran}
\usepackage{amsmath}
\usepackage[normalem]{ulem}
\usepackage{pgffor}
\usepackage{graphicx}
\usepackage{pgfplots}
\usepackage{caption}
\usepackage{float}
\usepackage{subcaption}
\usepackage{comment}
\usepackage{tikz}

\usepackage[colorlinks=true, allcolors=blue]{hyperref}
\usepackage{cite}

\usepackage{etoolbox}

\makeatletter
\patchcmd{\@maketitle}
  {\ifx\@empty\@thanks\else\thanks[\@empty]{\@thanks}\fi}
  {}{}{}
\makeatother

\title{Machine Learning-based Early Detection of Potato Sprouting Using Electrophysiological Signals}

\author{
\IEEEauthorblockN{
    Davide Andreoletti\IEEEauthorrefmark{1},
    Aris Marcolongo\IEEEauthorrefmark{2},
    Natasa Sarafijanovic Djukic\IEEEauthorrefmark{2},
    Julien Roulet\IEEEauthorrefmark{3},
    Stefano Billeter\IEEEauthorrefmark{4},
    Andrzej Kurenda\IEEEauthorrefmark{3},
    Margot Visse-Mansiaux\IEEEauthorrefmark{5},
    Brice Dupuis\IEEEauthorrefmark{5},
    Carrol Annette Plummer\IEEEauthorrefmark{3},
    Beatrice Paoli\IEEEauthorrefmark{2},
    Omran Ayoub\IEEEauthorrefmark{1}
}
\IEEEauthorblockA{\IEEEauthorrefmark{1}\thanks{Davide Andreoletti and Aris Marcolongo contributed equally to this work. This work has been supported by the Swiss Innovation Agency Innosuisse under Project \lq \lq Predicting Potato Sprouting to Optimise Tuber Storage (Project Nr. 100.494 IP-LS).}\textit{\small{Institute of Information Systems and Networking, University of Applied Sciences and Arts of Southern Switzerland, Lugano, Switzerland}}}\\
\IEEEauthorblockA{\IEEEauthorrefmark{2}\textit{\small{Laboratory for Web Science, Fernfachhochschule Schweiz, Brig, Switzerland,}}} \IEEEauthorblockA{\IEEEauthorrefmark{3}\textit{\small{Vivent SA, Gland, Switzerland}}}\\
\IEEEauthorblockA{\IEEEauthorrefmark{4}\textit{\small{University of Applied Sciences and Arts of Southern Switzerland, Lugano, Switzerland}}}\\
\IEEEauthorblockA{\IEEEauthorrefmark{5}\textit{\small{Agroscope, Swiss Confederation’s Center for Agricultural Research, Plant-Production Systems, Cultivation Techniques and Varieties in Arable Farming, Route de Duillier 60, 1260 Nyon, Switzerland}}}
}

\begin{document}
\maketitle

\begin{abstract}
Accurately predicting potato sprouting before the emergence of any visual signs is critical for effective storage management, as sprouting degrades both the commercial and nutritional value of tubers. Effective forecasting allows for the precise application of anti-sprouting chemicals (ASCs), minimizing waste and reducing costs. This need has become even more pressing following the ban on Isopropyl N-(3-chlorophenyl) carbamate (CIPC) or Chlorpropham due to health and environmental concerns, which has led to the adoption of significantly more expensive alternative ASCs. Existing approaches primarily rely on visual identification, which only detects sprouting after morphological changes have occurred, limiting their effectiveness for proactive management. A reliable early prediction method is therefore essential to enable timely intervention and improve the efficiency of post-harvest storage strategies, where early refers to detecting sprouting before any visible signs appear. In this work, we address the problem of early prediction of potato sprouting. To this end, we propose a novel machine learning (ML)-based approach that enables early prediction of potato sprouting using electrophysiological signals recorded from tubers using proprietary sensors. Our approach preprocesses the recorded signals, extracts relevant features from the wavelet domain, and trains supervised ML models for early sprouting detection. Additionally, we incorporate uncertainty quantification techniques to enhance predictions. Experimental results demonstrate promising performance in the early detection of potato sprouting by accurately predicting the exact day of sprouting for a subset of potatoes and while showing acceptable average error across all potatoes. Despite promising results, further refinements are necessary to minimize prediction errors, particularly in reducing the maximum observed deviations. 
By enabling reliable early detection, our solution has the potential to optimize storage strategies, minimize the use of expensive chemicals, and preserve tuber quality, thus delivering substantial added value to potato farmers and the wider potato supply chain.

\end{abstract}

\begin{IEEEkeywords}
Potato Sprouting Detection; Electrophysiological Signals; Wavelet Transform; Machine Learning; Uncertainty Quantification.
\end{IEEEkeywords}

\IEEEpeerreviewmaketitle

\section{Introduction}

Potatoes are the fourth most important food crop globally, and play a critical role in food security and economic sustainability \cite{Devaux2020, Devaux2014}. Given their significance, effective post-harvest storage strategies are essential to preserving their quality.

Typically, potatoes are stored for extended periods (e.g., up to 11 months) under controlled conditions to inhibit sprouting (i.e., germination). This is because sprouting leads to undesirable sugar accumulation in the tubers and weight loss, ultimately impacting their commercial and nutritional value. Therefore, mitigating sprouting during storage is critical for maintaining the quality and economic viability of tubers \cite{sugar}.

To mitigate sprouting, various anti-sprouting chemicals (ASCs) \cite{antisprouting, Paul2016} have been employed. Historically, Chlorpropham (CIPC) was widely used for this purpose; however, due to concerns over its potential health and environmental risks, it has been banned (CIPC has been banned in European and Swiss markets in 2020 \cite{Eu_Comm2019} \cite{Mahajan2008}). As an alternative, newer ASCs such as \lq\lq Dormir'' and \lq\lq Argos'' (orange oil) have been introduced. Despite their effectiveness, these alternatives are significantly more expensive than CIPC, costing multiple times more, and hence are not viable for application through a fixed calendar-based approach.

This shift highlights the growing need for precise and proactive sprouting prediction, enabling storage managers to apply ASCs only when necessary. By accurately forecasting the onset of sprouting, storage managers can optimize ASC application, reducing both economic costs and the environmental footprint associated with chemical treatments. 

Currently, sprouting detection methods primarily rely on visual inspection or imaging techniques, which depend on visible signs of sprouting (i.e., on morphological changes) 
\cite{Yu2015, Jin2015}, \cite{Qiao2005}, \cite{Visse-Mansiaux2022}. 
However, these approaches are unsuitable for predicting the occurrence of sprouting, as they can only detect sprouting once physical signs have already emerged (i.e., once sprouting has already occurred). This significantly reduces the effectiveness of these approaches for proactive storage management and commercial applications. Consequently, there is a pressing need for innovative solutions to predict sprouting occurrence prior to any visual signs. 

In this work, we address the challenge of early potato sprouting prediction by proposing a comprehensive machine learning-based approach that leverages electrophysiological signals recorded from potato tubers. To acquire these signals, we utilize for the first time proprietary sensors that measure the potential difference between set of probes connected to each of the potato tubers. Then, we employ signal processing and Machine Learning (ML) techniques to develop an end-to-end pipeline for an early prediction of potato sprouting. 

In terms of signal processing, we apply data preprocessing techniques and extract features from the wavelet domain to potentially capture the electrophysiological fingerprints associated with sprouting. Using extracted features, we train ML models in a supervised manner (we describe labeling process in detail in Sec. \ref{Sec:experiment}) for the task of early potato sprouting detection. Our pipeline also incorporates uncertainty quantification techniques to discard potentially uncertain predictions, with the aim of enhancing the final prediction of our pipeline. Note that, for the problem at hand, we are interested in one final prediction per potato tuber, as based on such a final prediction, the action of ASC spraying takes place. To this end, we build an approach that processes the predictions of the ML model, which are obtained at every instance $t$ for a given potato tuber, and estimates a sprouting day for the potato tuber. We test our approach on two datasets corresponding to potatoes stored at different temperatures. Each of the datasets consists of potato tubers of different varieties. Experimental results indicate that our proposed approach accurately predicts the exact sprouting day for a subset of potatoes and maintains a reasonable average error across all samples. However, further improvements are needed to reduce the maximum observed error. Additionally, incorporating uncertainty quantification enhances prediction consistency, lowering the mean absolute error and mitigating large deviations.

Our key contributions can be summarized as follows:
\begin{itemize}
    \item We design and conduct a series of controlled experiments to collect electrophysiological signals from stored potato tubers using proprietary sensors that measure the potential difference across probes connected to the potatoes. 
    \item We propose an end-to-end data pipeline that processes raw electrophysiological signals to predict potato sprouting. The pipeline includes data preprocessing techniques and ML models, enhanced with uncertainty quantification techniques, to capture distinct electrophysiological patterns indicative of sprouting. 
    \item We conduct experiments to obtain numerical results showcasing the effectiveness of our proposed approach for early detection of potato sprouting.  
\end{itemize}

The rest of the paper is organized as follows. Sec. \ref{relatedwork} discusses related work. Sec. \ref{Sec:experiment} describes the experimental setup for data collection and Sec. \ref{sec:methodology} introduces our methodology. Sec. \ref{evalsettings} describes the evaluation settings and Sec. \ref{sec:results} discusses experimental results. Finally, Sec. \ref{conclusion} concludes the paper. 

\begin{figure}[!t]
\centering
\subfloat[]{\includegraphics[width=2.5in, trim={0pt 80pt 00pt 220pt}, clip]{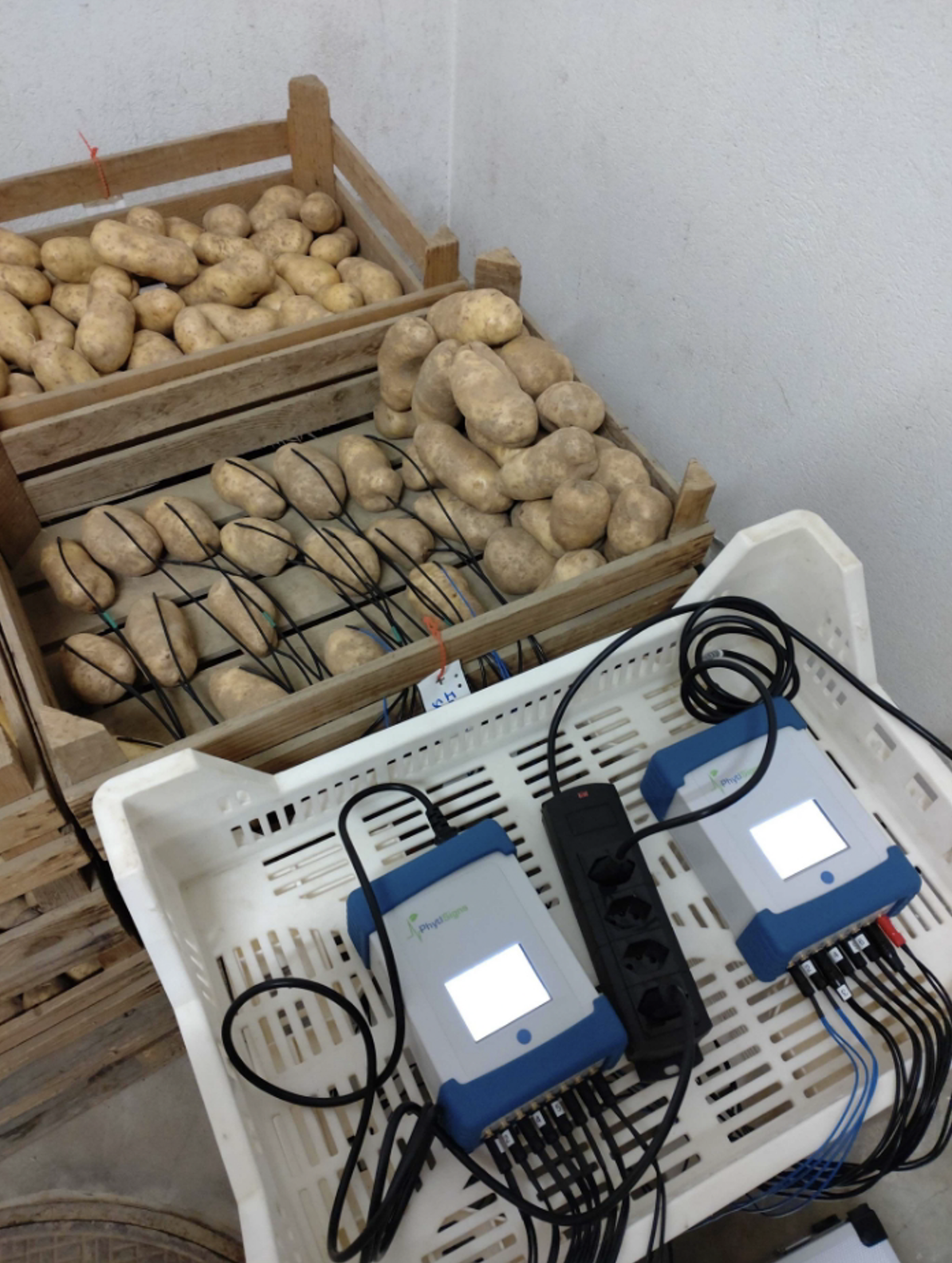}%
\label{fig:potatoes}}

\subfloat[]{\includegraphics[width=3.0in]{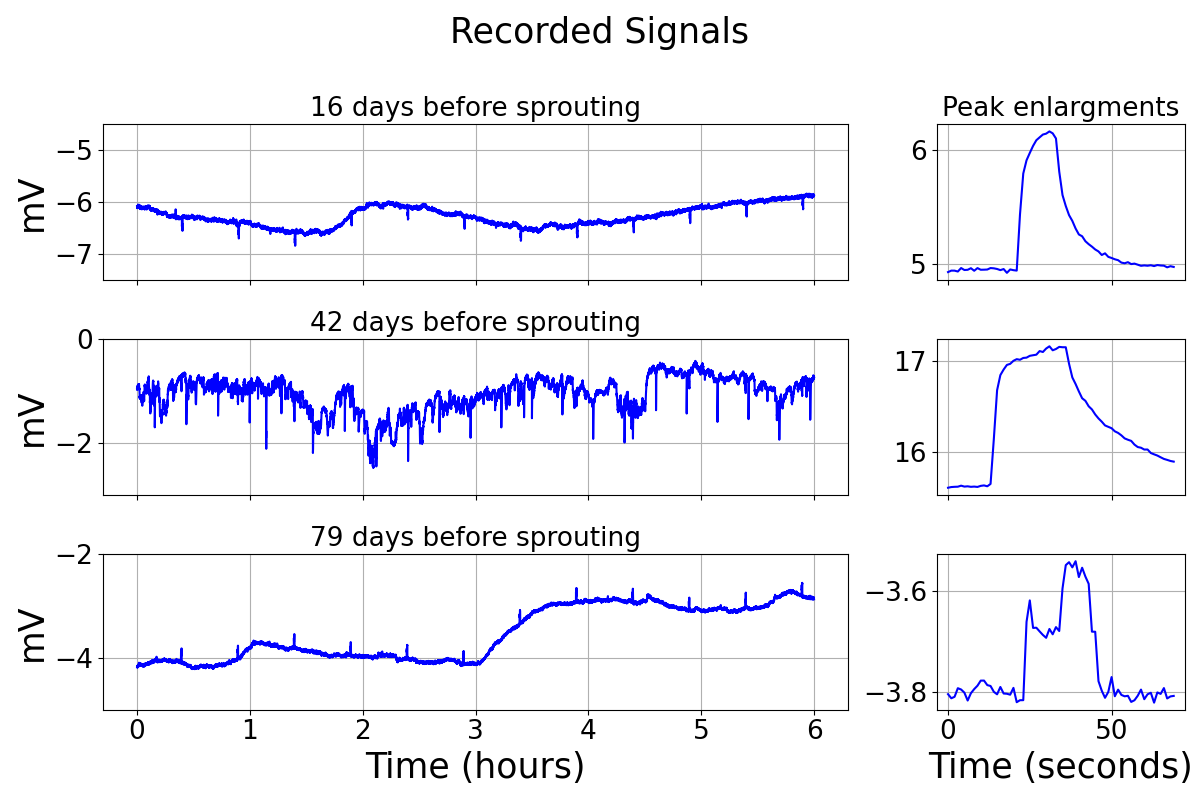}%
\label{fig:signals}}

\caption{(a) Photograph of the experimental setup, illustrating each sensor connected to its respective potato for recording electrophysiological activity.
(b) Examples of voltage signals recorded by the sensors over time, capturing electrophysiological activity. The signals exhibit a mix of low-frequency patterns intertwined with peaks associated with high-frequency activity, which are magnified on the right. The inherent variability in these signals makes it challenging to identify meaningful patterns through visual inspection alone. Consequently, a supervised machine learning model is required to extract and focus on patterns relevant for predicting the sprouting event.} 
\label{fig:experiment}
\end{figure}

\section{Related Work}\label{relatedwork}
The detection of potato sprouting has been extensively studied using statistical methods and machine learning techniques, including deep learning models. Most existing detection methods heavily rely on imaging techniques, utilizing hyperspectral imaging, machine vision, and deep neural networks to identify sprouting based on visual cues.

One of the earliest works in this domain, Qiao et al. \cite{Qiao2005}, proposed a methodology to estimate water content and weight using hyperspectral imaging integrated with an artificial neural network, achieving an accuracy of 97\% in sprouting identification. Similarly, Jin et al. \cite{Jin2015} employed hyperspectral imaging and demonstrated a slightly improved accuracy of 97.3\%, while Yu et al. \cite{Yu2015} adopted a support vector machine model trained on potato tuber images, achieving a 94\% identification rate. These early studies highlight the efficacy of hyperspectral imaging combined with machine learning for sprouting detection.

Further advancements introduced deep learning techniques to enhance detection accuracy and speed. In \cite{Yolov4}, the authors leveraged a lightweight image-based deep learning model to improve both the accuracy and computational efficiency of sprouting potato identification. Ming et al. \cite{Ming} presented an ensemble-based classifier approach using machine vision, which demonstrated robust performance in classifying sprouting stages. Ma et al. \cite{Ma} explored the use of convolutional neural networks (CNNs), specifically GoogLeNet, to detect potato sprouting with high accuracy. Wang et al. \cite{Wang} extended this approach by combining deep convolutional neural networks with transfer learning techniques to detect surface defects in potatoes, achieving an impressive 98.7\% accuracy. Rady et al. \cite{Rady2020} compared the capabilities of three different optical systems—visible/near-infrared (Vis/NIR) interactance spectroscopy, Vis/NIR hyperspectral imaging, and near-infrared (NIR) transmittance—combined with machine learning methods to detect sprouting activity. Their approach, which considered the primordial leaf count (LC) as a key feature, reached a detection accuracy of 90\%.

Despite the effectiveness of these approaches, they remain constrained to post-sprouting detection, focusing solely on identifying the occurrence of sprouting rather than predicting its onset. 

In our work, we address this limitation by proposing a novel approach that leverages electrophysiological signals recorded from potato tubers, combined with advanced signal processing and machine learning techniques, to enable early detection of potato sprouting. By early detection, we refer to identifying the onset of sprouting at least a week before any visible signs appear, which is particularly relevant given that ASC can be applied several days in advance of visible symptoms. Unlike existing methods that rely on visible signs, our approach aims to predict the onset of sprouting before any visual cues. 

A similar shift toward predictive modeling was recently explored by Visse-Mansiaux et al. \cite{Visse-Mansiaux2022}, who developed a model based on a comprehensive weather database containing 3,379 records from multi-year trials across more than 500 potato varieties. Their approach successfully forecasted dormancy end for potatoes stored at 8°C with a precision of 15 days, demonstrating the potential of environmental data-driven prediction for sprouting management. It is worth-noting that the use of plant electrophysiology with ML techniques has proven success for detection of spider mites in tomatoes \cite{spidermitestomatoes}, classification of nutrient deficiencies \cite{nutrientdeficienciestomatoes}, detection of abiotic stress \cite{Abioticstress} and detection of stress in tomatoes \cite{stresstomatoes}. Our work, however, is the first to explore the use of plant electrophysiology for the early prediction of sprouting in potato tubers. 

\section{Experimental Setup}\label{Sec:experiment}
The data collection process involves recording electrophysiological signals from stored potatoes by connecting sensor probes to the tubers, as illustrated in Fig. \ref{fig:potatoes}. 
For each potato tuber, the measurement setup consisted of 1-meter coaxial cables connected to two silver-plated needle electrodes. The reference electrode (50 mm long) was inserted into the center of the tuber, while the active electrode (5 mm long) was placed just beneath the surface in the vicinity of the apical bud. The electrical potential difference was continuously recorded at 256 Hz using Vivent Biosignals (Gland, Switzerland) PSR8 biosensor, an 8-channel amplifier with input impedance of 200 MOhm. After analogue-to-digital conversion, notch filters at 50 Hz and 100 Hz were applied to minimize mains interference. A biquad filter was then used before downsampling the signal to 1 Hz. Note that although the potatoes belonged to the same storage batch, sprouting does not occur simultaneously across all tubers. The onset of sprouting varies across individual potatoes, often spreading over several days or even weeks. Therefore, individual electrodes are necessary to capture tuber-specific electrophysiological changes. Nonetheless, detecting early sprouting in some tubers can provide valuable indications of broader physiological transitions within the batch.
 
In addition to capturing these signals, we employ video monitoring to track the status of each potato. The video monitoring serves to identify the ground truth, i.e., the exact date of sprouting. In other words, both the recording of electrophysiological signals and the video monitoring continues until a potato tuber has sprouted. Then, the recorded videos are analyzed to determine the exact sprouting time, i.e., for identifying the precise day on which sprouting occurs. 

The experiments involved three varieties of potatoes, namely Sorentina, SHC1010, and Agria, stored at 4 or 8 degrees, depending on potato variety. We purposely consider three varieties to enhance the robustness and generalizability of our findings across different potato varieties. Potatoes were not subject to any treatment to delay sprouting, e.g., spraying, other than the controlled temperature conditions. The recording process started at the moment of starting of potato storage and continued until sprouting occured, at which point data collection is terminated. The sprouting times ranged from 16th November 2023 to 8th March 2024 for potatoes stored at 8 degrees and from 28th December 2023 to 16th July 2024 for potatoes stored at 4 degrees.

\section{Methodology}\label{sec:methodology}

\begin{figure}[]
\centering
\includegraphics[width=3.3in, height=3.4cm]{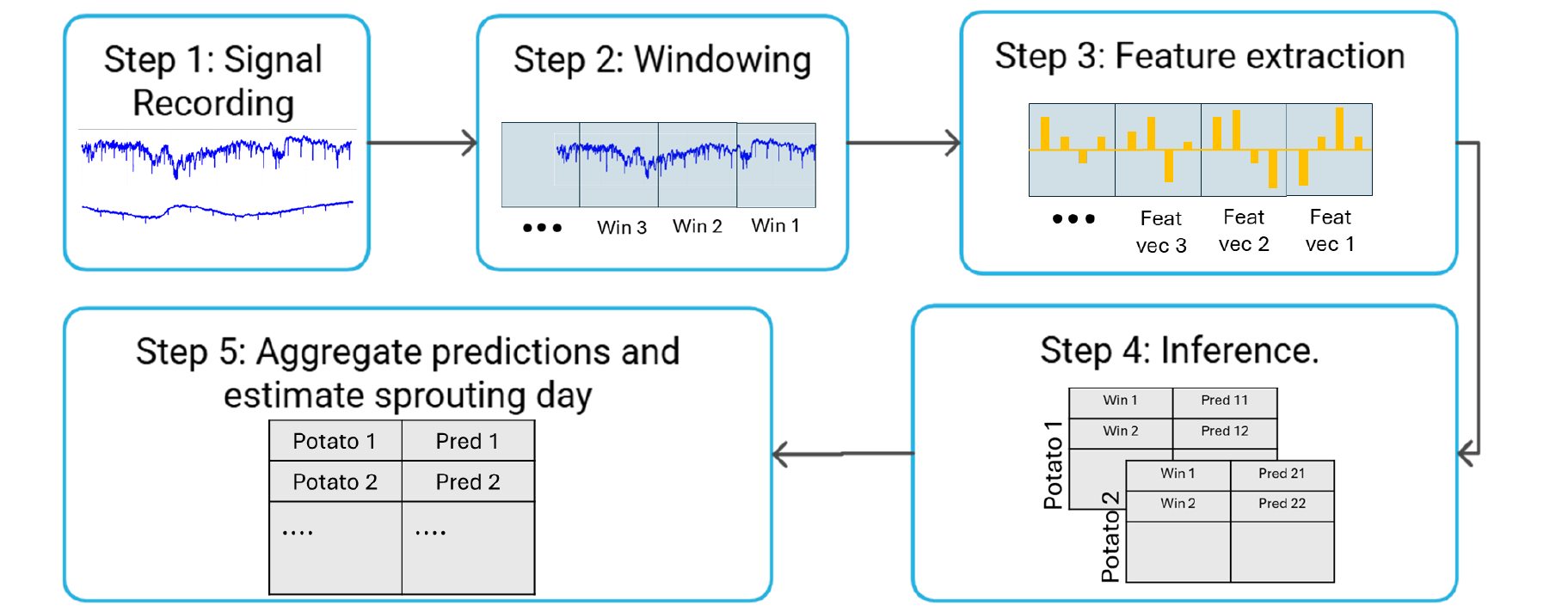}
\caption{Schematic representation of our proposed methodology. Electrophysiological signals recorded from sensors up to time $t$ for for each potato are initially segmented into windows (steps 1,2). Next, features are extracted and used to compute the estimated sprouting days $\hat D^{i}_{j}$ for each window $i$ of potato $j$ (steps 3,4). An aggregation procedure is then applied to derive the overall sprouting day estimate for each potato $\hat D_{j}$ (step 5).}
\label{fig:workflow}
\end{figure}

We formulate the problem of early potato sprouting detection as follows. Given the electrophysiological signal of a potato plant, our objective is to predict the day when sprouting will occur, referred to as the \emph{sprouting day}. More formally, we aim to learn an estimator that inputs the timeseries $\{s(t'),t' \le t\}$, which measures the electrophysiological activity of a plant up to the present time $t$, and estimates its future sprouting day $D$.

Our data pipeline consists of several steps involving signal preprocessing, feature extraction, ML modeling and Sprouting Day estimation, as shown in Fig. \ref{fig:workflow}. We describe these building blocks in the following subsections.

\subsection{Signal Preprocessing}
Given a set of \( N \) potato plants, we denote the time series of the \( j \)-th plant as \( s_{j}(t) \), \( \forall j \in [1,N] \). For clarity, we now describe the data preprocessing and feature extraction in detail, omitting the subscript \( j \) to improve readability.

The time series \( s(t) \) is first divided into \( M \) non-overlapping windows of size \( W \). Then, the continuous wavelet transform (CWT) \cite{wavelettransform} is computed for the \( i \)-th window \( w_i \), \( \forall i \in [1,M] \) for a set of \( K \) \textit{scales}. The concept of scale in the wavelet transform is analogous to frequency in Fourier analysis, but it refers to a range of frequencies rather than to a precise frequency. In this way, the wavelet transformation allows for a better balance between resolution in the time and frequency domains compared to alternative approaches such as the Short-Time Fourier Transform (STFT)\footnote{Our decision to leverage wavelet transformation for electrophysiological signals is driven by its proven effectiveness in similar data types, such as electrocardiogram (ECG) and electromyogram (EMG) signals \cite{wavelet1,wavelet2}.}. Notably, it is possible to determine the scale corresponding to a given frequency within the signal’s bandwidth. We select \( K \) scales, each corresponding to a specific frequency, which are equally spaced in the frequency domain on a logarithmic scale. This transformation yields a set of transformed windows $\mathcal{W}_{i}^{(k)}, \forall i \in [1,M], \forall k \in [1,K]$. $\mathcal{W}_{i}^{(k)}$ captures the evolution over time of the $i$-th window within the frequency range corresponding to the $k$-th scale. This multi-scale decomposition enables the identification of features at different temporal resolutions, making it particularly useful for analyzing signals with non-stationary properties, such as the considered electrophysiological signals.

Each transformed window \( \mathcal{W}_{i}^{(k)} \) then undergoes a feature extraction process to derive a compact yet informative representation of the signal. The extracted features include:

\begin{itemize}
    \item Energy: A measure of the signal's intensity, computed as the sum of squared values within the window.
    \item Statistical Descriptors: A set of robust statistical measures that summarize the distribution of values in the window, including the 5th, 25th, 75th, and 95th percentiles, median and mean values, standard deviation, minimum and maximum.
    \item Entropy: A measure of the uncertainty or randomness in the signal.
    \item Zero and Mean Crossings: The number of times the signal crosses zero and the mean value, which provides insight into the oscillatory nature of the signal.
\end{itemize}

As a result of this process, we obtain a feature vector \( \mathcal{V}_{i}^{(k)} \) for each window \( i \) and scale \( k \). Then, the feature vectors corresponding to different scales for the same window are concatenated, yielding a final representation \( F_{i} \) $ \equiv (\mathcal{V}_{i}^{(1)},...,\mathcal{V}_{i}^{(K)})$ for each window \( i \). We preproces the data of all the $N$ available plants, which results in \( F_{i}^{(j)} \) for each plant $j \in [1,N]$ and for each window $i \in [1,M_{j}]$, where $M_{j}$ is the number of windows of the $j$-th plant. 

\subsection{ML Model Development}\label{sec:ML_model}
The feature vector for the \( i \)-th window of potato $j$, \( F_{i}^{(j)} \), is associated with the target \( Y_{i}^{(j)} \). The latter corresponds to the number of days between the recording day of the window considered and \( D_{j} \), the actual sprouting day of the \( j \)-th plant. We construct the dataset consisting of features \( \mathcal{X} = \{ F_{i}^{(j)} \} \) and targets \( \mathcal{Y} = \{ Y_{i}^{(j)} \} \), with \( j \in [1,N] \) and \( i \in [1,M_{j}] \), which serve to train a supervised regression model. Using the training dataset, the model learns to estimate the number of days remaining until the sprouting event based on the features extracted from each plant's window.

We perform training using two main strategies. In the first approach, referred to as \textit{Single Model}, the entire training dataset is used to train a single estimator. In the second approach, referred to as \textit{Multiple Models with Uncertainty Quantification}, the training dataset is randomly split into 10 subsets of equal size, each used to train a separate estimator. Having multiple estimates allows us to quantify the agreement among models in making predictions. This, in turn, helps to filter out predictions where the model exhibits a certain level of uncertainty. We formalize the training procedure for both the single-estimator case and the multiple-estimator approach with uncertainty quantification (UQ) in the next subsection.

\subsection{Sprouting Day Estimation}\label{sec:SD_estimation}

\subsubsection{Single Model}

After the training phase, we obtain a regressor that, given an unseen vector \( F_{i}^{(j)} \) as input, produces \( \hat{Y}_{i}^{(j)} \), i.e., the estimated number of days between the \( i \)-th window of the \( j \)-th plant and its sprouting day \( D_{j} \). 

Hence, by querying the trained machine learning model for all the windows of the \( j \)-th plant, we obtain a set of estimated days until sprouting: 

\[
\{ \hat{Y}_{i}^{(j)} \}, \quad \forall i \in [1,M_{j}].
\]

At this point, we compute a set of estimated sprouting days by adding \( \hat{Y}_{i}^{(j)} \) to the day corresponding to the \( i \)-th window, resulting in:  

\[
\{ \hat{D}_{i}^{(j)} \}, \quad \hat{D}_{i}^{(j)} = d_{i}^{(j)} + \hat{Y}_{i}^{(j)}, \quad \forall i \in [1,M_{j}].
\]

where \( d_{i}^{(j)} \) is the day to which the \( i \)-th window belongs.

To obtain the final estimated sprouting day for the \( j \)-th plant, we compute the average over all estimated sprouting days. Formally:

\[
\hat{D}_{j} = \frac{1}{M^{(t)}_{j}} \sum_{i=1}^{M^{(t)}_{j}} \hat{D}_{i}^{(j)}.
\]

, where $M^{(t)}_{j}$ is the number of windows of the $j$-th plant before $t$, which is the time instant when the utilizer of the model is supposed to stop observing the data to compute the estimated sprouting day. It is worth noting that more sophisticated approaches beyond simple averaging could be explored, which we leave as future work.

\subsubsection{Multiple Models with Uncertainty Quantification (UQ)}

After the training phase, we obtain 10 regressors. Each \( u \)-th regressor, \( \forall u \in \{1,10\} \), is queried on the unseen vector \( F_{i}^{(j)} \) and produces \( \hat{Y}_{iu}^{(j)} \), representing the estimated number of days between the \( i \)-th window of the \( j \)-th plant and its sprouting day \( D_{j} \), according to the \( u \)-th estimator.  For UQ, the notation \( \hat{Y}_{i}^{(j)} \) (without the index $u$) refers to the mean over the 10 predictions. The same set of estimates can be used to quantify the uncertainty in the mean prediction. Specifically, we compute the 95\% confidence interval from this population of predictions. If the confidence interval exceeds a predefined threshold \( UQ_{th} \), a parameter to be tuned, the window is discarded. This results in a set of windows for the \( j \)-th plant, denoted as $R_{j}$ which are considered for further processing. Than, the formulas for the data processing pipeline in the previous section apply almost directly, with summations restricted to the set of non-discarded windows $R_j$. 






\section{Evaluation Settings}\label{evalsettings}

\subsection{Dataset Description}
The dataset at 8 degrees consisted of 16, 16, and 32 potatoes for Sorentina, SHC1010, and Agria, respectively. At 4 degrees, the dataset included 23, 27, and 14 potatoes for the same varieties. These two datasets, referred to as Dataset 1 (8°C) and Dataset 2 (4°C), contain electrophysiological signals recorded over a period ranging from a minimum of 1 month to a maximum of 9 months.

\subsection{Model Training and Testing}
We consider XGBoost as ML model. To train and evaluate our models, we adopt a leave-one-out cross-validation (LOO-CV) approach and report the average performance across all evaluations. Specifically, given a dataset of $N$ potatoes, we train $N$ distinct models, each time excluding one potato from the training set. The excluded potato serves as the test sample for that iteration. This process ensures that no individual potato has its electrophysiological data simultaneously present in both the training and test sets of any model, effectively preventing information leakage and providing a robust assessment of the model’s generalization capability.

\subsection{Evaluation Metrics}

We evaluate the performance of the estimators using two main metrics: the Mean Absolute Error (MAE) and the Error in Sprouting Day (ESD). More explicitly:

\begin{equation}
MAE=\frac{1}{N}\sum_{j=1}^N \frac{1}{M_j} \sum_{i=1}^{M_j} |\hat{Y}_{i}^{(j)}-Y_{i}^{j}| \nonumber
\end{equation}

\begin{align}
&MAE=\frac{1}{N}\sum_{j=1}^N MAE_j, \nonumber \\ &MAE_j=\frac{1}{M_j} \sum_{i=1}^{M_j} |\hat{Y}_{i}^{(j)}-Y_{i}^{j}| \nonumber
\end{align}
and:
\begin{equation}
ESD=\frac{1}{N}\sum_{j=1}^N ESD_j,  ESD_j = |\hat{D}_j-D_{j}|, \nonumber
\end{equation}

where the same notation of the previous section was used. The MAE quantifies the typical error of each per-window prediction, whereas the ESD estimates the typical error per-potato obtained after performing the averaging of predictions over windows.




\subsection{Model Validation}

We inspect the quality of per-window predictions by looking closely at statistical relations between the exact target value $Y$ and the estimated days to sprouting $\hat Y$. In particular we monitor the expectation $E[Y|\hat Y]$ and the variance $Var(Y|\hat Y)$ (or $Std(Y|\hat Y)$) as a function of $\hat Y$ values. The first expression corresponds to the mean of all exact values from samples sharing the same predicted value $\hat Y$, whereas the second expression evaluates the variance (or the standard deviation) of these exact values. By analogy with the terminology often used in classification, we refer to the curve of $E[Y|\hat Y]$ as a function of $\hat Y$ values as the \emph{calibration curve}, indicating whether, on average, the predictions are systematically overestimating or underestimating the true values. Instead, an analysis of the conditional variance tells us how much we can trust the individual predictions. Moreover, this validation setting leads to some desirable statistical properties which help model understanding. First, a model trained with infinite data under the mean-squared error respects the relation $E[Y|\hat Y]=\hat Y$. Therefore, by analyzing departure from this ideal behavior we get an indication about the overall quality of the training procedure, given the features used by the model. Second, the low of total variance reads in this context $Var(Y)=Var(\hat Y)+E[Var(Y|\hat Y)]$, showing that models based on features with high predictive power and explained variance will have a low value of $Var(Y|\hat Y)$. Therefore, by analyzing the magnitude of $Var(Y|\hat Y)$ we get a grasp of the feature quality. 

\section{Results}\label{sec:results}

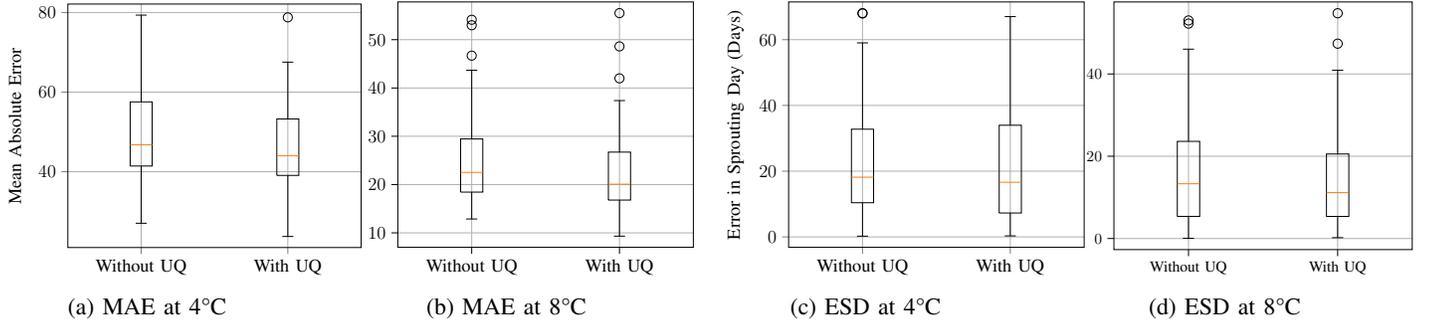
\begin{figure*}
    \centering
    \begin{subfigure}{0.21\textwidth}
        \centering
        \resizebox{!}{3.7cm}{
\begin{tikzpicture}

\definecolor{darkgray176}{RGB}{176,176,176}
\definecolor{darkorange25512714}{RGB}{255,127,14}

\begin{axis}[
tick align=outside,
tick pos=left,
x grid style={darkgray176},
xlabel={},
xlabel style={font=\large}, 
xmajorgrids,
xmin=0.5, xmax=2.5,
xtick={1,2},
xticklabels={Without UQ,With UQ},
ticklabel style={font=\large}, %
y grid style={darkgray176},
ylabel={Mean Absolute Error},
ylabel style={font=\large}, 
ymajorgrids,
ymin=20.9447835829193, ymax=82.1198310529559,
legend style={font=\large}
]
\addplot [black]
table {%
0.925 41.4269856915173
1.075 41.4269856915173
1.075 57.5441480490818
0.925 57.5441480490818
0.925 41.4269856915173
};
\addplot [black]
table {%
1 41.4269856915173
1 27.0477862603524
};
\addplot [black]
table {%
1 57.5441480490818
1 79.3391470770451
};
\addplot [black]
table {%
0.9625 27.0477862603524
1.0375 27.0477862603524
};
\addplot [black]
table {%
0.9625 79.3391470770451
1.0375 79.3391470770451
};
\addplot [black]
table {%
1.925 39.0686849134575
2.075 39.0686849134575
2.075 53.2541439955728
1.925 53.2541439955728
1.925 39.0686849134575
};
\addplot [black]
table {%
2 39.0686849134575
2 23.72546755883
};
\addplot [black]
table {%
2 53.2541439955728
2 67.5274231055264
};
\addplot [black]
table {%
1.9625 23.72546755883
2.0375 23.72546755883
};
\addplot [black]
table {%
1.9625 67.5274231055264
2.0375 67.5274231055264
};
\addplot [black, mark=o, mark size=3, mark options={solid,fill opacity=0}, only marks]
table {%
2 78.7698831677603
};
\addplot [darkorange25512714]
table {%
0.925 46.7607707365136
1.075 46.7607707365136
};
\addplot [darkorange25512714]
table {%
1.925 44.0182004849603
2.075 44.0182004849603
};
\end{axis}

\end{tikzpicture}}
        \caption{MAE at 4°C}
        \label{fig:subfig3}
    \end{subfigure}
    \hfill
    \begin{subfigure}{0.21\textwidth}
        \centering
        \resizebox{!}{3.7cm}{
\begin{tikzpicture}

\definecolor{darkgray176}{RGB}{176,176,176}
\definecolor{darkorange25512714}{RGB}{255,127,14}

\begin{axis}[
tick align=outside,
tick pos=left,
x grid style={darkgray176},
xlabel={},
xlabel style={font=\large}, 
xmajorgrids,
xmin=0.5, xmax=2.5,
xtick style={color=black},
xtick={1,2},
xticklabels={Without UQ,With UQ},
ticklabel style={font=\large}, %
y grid style={darkgray176},
ymajorgrids,
ymin=7.01372604168068, ymax=57.8096966426027,
ytick style={color=black},
legend style={font=\large} 
]
\addplot [black]
table {%
0.925 18.4404553272419
1.075 18.4404553272419
1.075 29.4476676668522
0.925 29.4476676668522
0.925 18.4404553272419
};
\addplot [black]
table {%
1 18.4404553272419
1 12.8711151298212
};
\addplot [black]
table {%
1 29.4476676668522
1 43.6790059079829
};
\addplot [black]
table {%
0.9625 12.8711151298212
1.0375 12.8711151298212
};
\addplot [black]
table {%
0.9625 43.6790059079829
1.0375 43.6790059079829
};
\addplot [black, mark=o, mark size=3, mark options={solid,fill opacity=0}, only marks]
table {%
1 52.9919925853967
1 46.6935992299775
1 54.1229421940256
};
\addplot [black]
table {%
1.925 16.7891628346919
2.075 16.7891628346919
2.075 26.7188619510302
1.925 26.7188619510302
1.925 16.7891628346919
};
\addplot [black]
table {%
2 16.7891628346919
2 9.32263379626804
};
\addplot [black]
table {%
2 26.7188619510302
2 37.3908256693474
};
\addplot [black]
table {%
1.9625 9.32263379626804
2.0375 9.32263379626804
};
\addplot [black]
table {%
1.9625 37.3908256693474
2.0375 37.3908256693474
};
\addplot [black, mark=o, mark size=3, mark options={solid,fill opacity=0}, only marks]
table {%
2 41.994569174851
2 48.6080263822507
2 55.5007888880153
};
\addplot [darkorange25512714]
table {%
0.925 22.5216786946832
1.075 22.5216786946832
};
\addplot [darkorange25512714]
table {%
1.925 20.0719045318737
2.075 20.0719045318737
};
\end{axis}

\end{tikzpicture}}
        \caption{MAE at 8°C}
        \label{fig:subfig4}
    \end{subfigure}
    \hfill
    \begin{subfigure}{0.21\textwidth}
        \centering
        \resizebox{!}{3.7cm}{
\begin{tikzpicture}

\definecolor{darkgray176}{RGB}{176,176,176}
\definecolor{darkorange25512714}{RGB}{255,127,14}

\begin{axis}[
tick align=outside,
tick pos=left,
x grid style={darkgray176},
xlabel={},
xlabel style={font=\large}, 
xmajorgrids,
xmin=0.5, xmax=2.5,
xtick={1,2},
xticklabels={Without UQ,With UQ},
ticklabel style={font=\large}, %
y grid style={darkgray176},
ylabel={Error in Sprouting Day (Days)},
ylabel style={font=\large}, 
ymajorgrids,
ymin=-3.15649167065888, ymax=71.4729095232991,
legend style={font=\large} 
]
\addplot [black]
table {%
0.925 10.4007175500247
1.075 10.4007175500247
1.075 32.7686402378584
0.925 32.7686402378584
0.925 10.4007175500247
};
\addplot [black]
table {%
1 10.4007175500247
1 0.235753838157393
};
\addplot [black]
table {%
1 32.7686402378584
1 59.0266976752436
};
\addplot [black]
table {%
0.9625 0.235753838157393
1.0375 0.235753838157393
};
\addplot [black]
table {%
0.9625 59.0266976752436
1.0375 59.0266976752436
};
\addplot [black, mark=o, mark size=3, mark options={solid,fill opacity=0}, only marks]
table {%
1 68.0806640144828
1 67.9650740180633
};
\addplot [black]
table {%
1.925 7.2737369785531
2.075 7.2737369785531
2.075 33.9739535215012
1.925 33.9739535215012
1.925 7.2737369785531
};
\addplot [black]
table {%
2 7.2737369785531
2 0.311669540405262
};
\addplot [black]
table {%
2 33.9739535215012
2 67.0166934917925
};
\addplot [black]
table {%
1.9625 0.311669540405262
2.0375 0.311669540405262
};
\addplot [black]
table {%
1.9625 67.0166934917925
2.0375 67.0166934917925
};
\addplot [darkorange25512714]
table {%
0.925 18.1692261669711
1.075 18.1692261669711
};
\addplot [darkorange25512714]
table {%
1.925 16.6251455414498
2.075 16.6251455414498
};
\end{axis}

\end{tikzpicture}}
        \caption{ESD at 4°C}
        \label{fig:subfig1}
    \end{subfigure}
    \hfill
    \begin{subfigure}{0.21\textwidth}
        \centering
        \resizebox{!}{3.7cm}{
\begin{tikzpicture}

\definecolor{darkgray176}{RGB}{176,176,176}
\definecolor{darkorange25512714}{RGB}{255,127,14}

\begin{axis}[
tick align=outside,
tick pos=left,
x grid style={darkgray176},
xlabel={},
xlabel style={font=\large}, 
xmajorgrids,
xmin=0.5, xmax=2.5,
xtick style={color=black},
xtick={1,2},
xticklabels={Without UQ,With UQ},
y grid style={darkgray176},
ymajorgrids,
ymin=-2.69817934536226, ymax=57.4968118574282,
ytick style={color=black}
]
\addplot [black]
table {%
0.925 5.37845591385439
1.075 5.37845591385439
1.075 23.6137557894545
0.925 23.6137557894545
0.925 5.37845591385439
};
\addplot [black]
table {%
1 5.37845591385439
1 0.0379566184009406
};
\addplot [black]
table {%
1 23.6137557894545
1 46.0264058054229
};
\addplot [black]
table {%
0.9625 0.0379566184009406
1.0375 0.0379566184009406
};
\addplot [black]
table {%
0.9625 46.0264058054229
1.0375 46.0264058054229
};
\addplot [black, mark=o, mark size=3, mark options={solid,fill opacity=0}, only marks]
table {%
1 52.2455811021431
1 53.0047074243532
};
\addplot [black]
table {%
1.925 5.36883747580102
2.075 5.36883747580102
2.075 20.5442816621063
1.925 20.5442816621063
1.925 5.36883747580102
};
\addplot [black]
table {%
2 5.36883747580102
2 0.204116588378497
};
\addplot [black]
table {%
2 20.5442816621063
2 40.9202556851544
};
\addplot [black]
table {%
1.9625 0.204116588378497
2.0375 0.204116588378497
};
\addplot [black]
table {%
1.9625 40.9202556851544
2.0375 40.9202556851544
};
\addplot [black, mark=o, mark size=3, mark options={solid,fill opacity=0}, only marks]
table {%
2 47.3632292869763
2 54.760675893665
};
\addplot [darkorange25512714]
table {%
0.925 13.3095080777042
1.075 13.3095080777042
};
\addplot [darkorange25512714]
table {%
1.925 11.1534264693999
2.075 11.1534264693999
};
\end{axis}

\end{tikzpicture}}
        \caption{ESD at 8°C}
        \label{fig:subfig2}
    \end{subfigure}

    \caption{Comparison of Mean Absolute Error and Error in Sprouting Days for Datasets 1 and 2, comprising potatoes stored at 4°C and 8°C, respectively, with and without Uncertainty Quantification.}
    \label{fig:maesprouting}
\end{figure*}

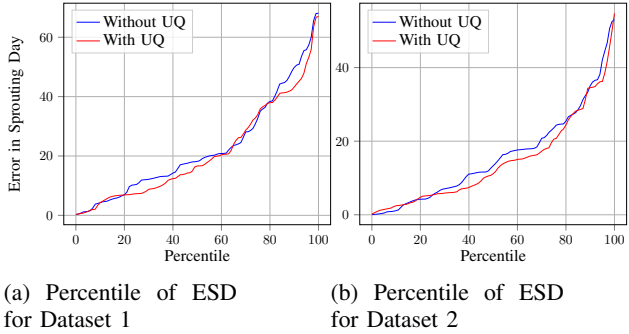
\begin{figure}
    \begin{subfigure}{0.35\linewidth} 
        \resizebox{!}{3.5cm}{
\begin{tikzpicture}

\definecolor{darkgray176}{RGB}{176,176,176}
\definecolor{lightgray204}{RGB}{204,204,204}

\begin{axis}[
legend cell align={left},
legend style={
  fill opacity=0.8,
  draw opacity=1,
  text opacity=1,
  at={(0.03,0.97)},
  anchor=north west,
  draw=lightgray204
},
tick align=outside,
tick pos=left,
x grid style={darkgray176},
xlabel={Percentile},
xlabel style={font=\large}, 
xmajorgrids,
xmin=-5, xmax=105,
y grid style={darkgray176},
ylabel={Error in Sprouting Day},
ylabel style={font=\large}, 
ymajorgrids,
ymin=-3.15649167065888, ymax=71.4729095232991,
legend style={font=\large} 
]
\addplot [semithick, blue]
table {%
0 0.235753774642944
1 0.442328691482544
2 0.730570673942566
3 1.1350314617157
5 1.33332133293152
6 1.60516130924225
7 2.5476438999176
8 3.78146266937256
9 4.00460433959961
10 4.26552724838257
11 4.56800985336304
12 4.6585111618042
13 4.81409406661987
14 5.18175411224365
15 5.46783065795898
16 5.70924425125122
17 5.86794805526733
18 6.1731972694397
19 6.60344123840332
20 6.91933631896973
21 7.79421615600586
22 9.6511869430542
23 10.150089263916
25 10.4007177352905
26 10.9941549301147
27 11.8764524459839
30 12.1420392990112
32 12.4862232208252
34 12.9126205444336
35 13.0923051834106
37 13.1745386123657
38 13.2314643859863
40 14.2337026596069
41 14.5940208435059
42 15.7518243789673
43 17.0257434844971
45 17.402889251709
46 17.5558490753174
47 17.817590713501
48 18.0004482269287
49 18.0493202209473
50 18.169225692749
51 18.411003112793
52 19.0504894256592
53 19.4776859283447
54 19.7722702026367
55 20.0065841674805
56 20.1693286895752
57 20.2426090240479
58 20.5462989807129
59 20.8059234619141
61 20.867130279541
62 20.9859256744385
63 22.0501480102539
64 22.8868198394775
65 23.5030536651611
66 23.7666339874268
67 24.0684833526611
68 24.5076656341553
69 26.255485534668
70 28.1098346710205
71 28.167688369751
72 28.5912227630615
73 29.2890129089355
74 30.8724784851074
75 32.7686386108398
76 35.1183853149414
77 35.8219757080078
78 36.3808670043945
79 37.788990020752
80 38.3705711364746
81 38.546314239502
82 40.1105651855469
83 41.9868125915527
84 44.2288513183594
85 44.5269355773926
86 44.7388534545898
87 45.442253112793
88 46.848575592041
89 48.5147438049316
90 49.8314094543457
91 50.6130828857422
92 50.9084014892578
93 53.5500030517578
94 55.4847640991211
95 55.8057861328125
96 57.3915672302246
97 60.0099182128906
98 65.6410980224609
99 68.0078430175781
100 68.0806655883789
};
\addlegendentry{Without UQ}
\addplot [semithick, red]
table {%
0 0.311669588088989
3 0.766777753829956
4 1.01296055316925
5 1.3215526342392
6 1.76062023639679
7 1.95191419124603
8 2.08877182006836
9 3.3837308883667
10 4.28200960159302
11 4.74394035339355
13 5.72642993927002
14 6.13660717010498
15 6.38548469543457
16 6.55934047698975
18 6.75144052505493
20 6.8899998664856
22 6.97855091094971
24 7.22763442993164
27 7.39444923400879
28 7.69784879684448
29 8.13989734649658
30 8.76681232452393
31 8.91132259368896
32 9.01088523864746
33 9.24578666687012
34 9.54720020294189
35 9.88902187347412
36 10.3138017654419
37 10.9461097717285
38 11.7926387786865
39 12.1351938247681
40 12.3634920120239
41 12.4601974487305
42 13.0309391021729
43 13.6762409210205
44 13.7177820205688
45 13.9248886108398
46 14.2644491195679
47 14.3479089736938
48 14.9229736328125
49 16.3105392456055
50 16.6251449584961
52 16.6975326538086
53 17.1165580749512
54 17.7677326202393
55 18.3208198547363
56 18.9825057983398
57 19.7799434661865
59 20.0390434265137
60 20.4191379547119
61 20.5602264404297
62 20.6017475128174
63 20.7534446716309
64 21.8670806884766
65 23.9125938415527
66 25.2574272155762
67 26.1473999023438
68 26.2484588623047
69 27.3421592712402
70 28.6838836669922
71 29.5491580963135
72 30.712869644165
73 32.1004104614258
74 32.8012199401855
75 33.9739532470703
76 35.8808555603027
77 36.5751914978027
78 37.0429725646973
79 37.7163887023926
80 37.9654769897461
81 38.0084800720215
82 38.8096237182617
83 39.8226280212402
84 41.0840263366699
85 41.2498893737793
86 41.3128280639648
87 41.5168380737305
88 41.8340339660645
89 42.269718170166
91 44.2306976318359
93 46.1322212219238
94 47.8957405090332
95 51.164722442627
96 53.2106056213379
97 55.9027633666992
98 63.4568901062012
99 66.7380905151367
100 67.0166931152344
};
\addlegendentry{With UQ}
\end{axis}

\end{tikzpicture}}
        \caption{Percentile of ESD for Dataset 1}
        \label{fig:subfig6}
    \end{subfigure}
    \hspace{1cm}
    \begin{subfigure}{0.35\linewidth} 
        \resizebox{!}{3.5cm}{
\begin{tikzpicture}

\definecolor{darkgray176}{RGB}{176,176,176}
\definecolor{lightgray204}{RGB}{204,204,204}

\begin{axis}[
legend cell align={left},
legend style={
  fill opacity=0.8,
  draw opacity=1,
  text opacity=1,
  at={(0.03,0.97)},
  anchor=north west,
  draw=lightgray204
},
tick align=outside,
tick pos=left,
x grid style={darkgray176},
xlabel={Percentile},
xlabel style={font=\large}, 
xmajorgrids,
xmin=-5, xmax=105,
y grid style={darkgray176},
ymajorgrids,
ymin=-2.69817934536226, ymax=57.4968118574282,
legend style={font=\large} 
]
\addplot [semithick, blue]
table {%
0 0.0379565954208374
2 0.162778854370117
4 0.343001842498779
5 0.473186016082764
6 0.745710611343384
7 0.859549522399902
8 0.891872406005859
9 0.977776527404785
10 1.12793672084808
11 1.34877848625183
12 2.05789494514465
13 2.69156432151794
14 3.00917029380798
16 3.55267262458801
17 3.82015013694763
18 4.03078842163086
19 4.192946434021
21 4.25805139541626
22 4.25823259353638
23 4.45270967483521
24 4.77082252502441
25 5.37845611572266
26 5.81961584091187
27 6.1533145904541
28 6.61479663848877
29 6.93210983276367
32 7.32755327224731
35 7.80938625335693
36 8.17870235443115
37 8.77984809875488
39 10.4699792861938
40 11.0687952041626
41 11.1261711120605
44 11.5469274520874
45 11.6006164550781
47 11.6243209838867
48 11.9180278778076
49 12.6560277938843
51 13.9653596878052
52 14.7148551940918
54 16.3511333465576
55 16.4208908081055
56 16.6746444702148
57 17.158390045166
58 17.3380126953125
60 17.6106033325195
61 17.6854820251465
62 17.7351760864258
63 17.8356647491455
64 17.8948631286621
66 17.9553031921387
67 18.0933589935303
68 18.4212322235107
69 19.5601329803467
70 20.7784156799316
71 20.954122543335
72 21.5257263183594
73 22.3942527770996
74 22.9748954772949
75 23.6137561798096
76 24.3481616973877
77 24.5614471435547
78 24.6423664093018
79 24.6891574859619
80 25.4724674224854
81 26.6386070251465
83 27.3657455444336
84 27.7656536102295
85 28.5890045166016
86 29.6817054748535
87 31.2938060760498
88 32.3857192993164
89 33.3470840454102
90 35.0610618591309
91 36.1031951904297
92 36.5345573425293
93 36.712776184082
94 38.1773681640625
95 42.0712852478027
96 44.451847076416
97 46.7105140686035
98 50.6285972595215
99 52.5264587402344
100 53.0047073364258
};
\addlegendentry{Without UQ}
\addplot [semithick, red]
table {%
0 0.204116582870483
2 0.849361419677734
3 1.13105070590973
4 1.30977547168732
5 1.45581603050232
7 1.70231020450592
8 1.86498725414276
9 2.2393274307251
10 2.46317768096924
11 2.52148914337158
13 2.72630214691162
14 2.80605506896973
15 3.03534197807312
16 3.30453729629517
17 3.43688225746155
18 3.79011607170105
19 4.3317551612854
20 4.77693033218384
21 5.07167053222656
22 5.11317205429077
24 5.27654457092285
25 5.36883735656738
26 5.52937984466553
27 5.73161697387695
28 5.73506736755371
29 5.78714084625244
30 5.90404510498047
32 6.01456212997437
33 6.04149913787842
34 6.11442089080811
35 6.24269008636475
36 6.74627590179443
37 7.05128240585327
39 7.24120235443115
40 7.40806293487549
41 7.74268341064453
43 8.25885200500488
44 8.6182804107666
45 9.20145702362061
46 9.96363258361816
47 10.2914981842041
48 10.546142578125
49 10.7049407958984
50 11.1534261703491
51 11.7644529342651
52 12.7109460830688
53 13.3626222610474
54 13.8289699554443
55 14.1762609481812
56 14.4633703231812
57 14.6752510070801
59 14.8369016647339
60 15.0439434051514
61 15.1233377456665
62 15.1615800857544
63 15.3728847503662
64 15.6258106231689
65 15.9190559387207
68 16.293420791626
69 16.7415447235107
70 17.282995223999
71 17.739896774292
72 18.0342292785645
73 18.2066345214844
74 19.60036277771
75 20.5442810058594
76 20.7745475769043
77 21.7391586303711
78 22.7496395111084
79 23.3158988952637
80 24.4228324890137
81 25.8174819946289
82 26.7487182617188
83 27.5399112701416
84 28.1669139862061
85 28.410961151123
86 28.6178150177002
87 28.8709468841553
88 31.3300266265869
89 34.3772315979004
91 34.6650428771973
92 34.8435211181641
93 35.6094818115234
94 36.1523971557617
95 36.2053985595703
96 38.4750900268555
97 41.6289825439453
98 45.6880569458008
99 50.100284576416
100 54.7606773376465
};
\addlegendentry{With UQ}
\end{axis}

\end{tikzpicture}}
        \caption{Percentile of ESD for Dataset 2}
        \label{fig:subfig7}
    \end{subfigure}

    \caption{Percentile curve of ESD for Dataset 1 and Dataset 2.}
    \label{fig:percentile_comparison}
\end{figure}

\begin{figure}[b]
    \begin{subfigure}{0.35\linewidth} 
        \resizebox{!}{3.5cm}{
\begin{tikzpicture}

\definecolor{darkgray176}{RGB}{176,176,176}
\definecolor{darkorange25512714}{RGB}{255,127,14}
\definecolor{lightgray204}{RGB}{204,204,204}
\definecolor{steelblue31119180}{RGB}{31,119,180}

\begin{axis}[
legend cell align={left},
legend style={fill opacity=0.8, draw opacity=1, text opacity=1, draw=lightgray204},
tick align=outside,
tick pos=left,
x grid style={darkgray176},
y grid style={darkgray176},
xmajorgrids,
ymajorgrids,
xlabel={$t_{lag}$},
xlabel style={font=\large}, 
ylabel={$ESD$},
ylabel style={font=\large}, 
xmin=-30.45, xmax=1.45,
xtick style={color=black},
y grid style={darkgray176},
ymin=20.7895079070547, ymax=24.9658725542486,
ytick style={color=black},
legend style={font=\large} 
]
\addplot [semithick, steelblue31119180]
table {%
-29 24.776037797558
-28 24.6280599712173
-27 24.4954454327939
-26 24.3545565152989
-25 24.233362525374
-24 24.1160812084269
-23 23.9962499277918
-22 23.8654748607835
-21 23.7480846085897
-20 23.6434483380272
-19 23.5281330291761
-18 23.407167876707
-17 23.3136185428255
-16 23.2277025436619
-15 23.1243936828416
-14 23.0314955134501
-13 22.9306014374693
-12 22.8523886212598
-11 22.7733102655701
-10 22.728261731263
-9 22.7108185587526
-8 22.6897695257331
-7 22.6610932658577
-6 22.6558109789828
-5 22.6544072277714
-4 22.6594691439253
-3 22.6567930194711
-2 22.6726757880462
-1 22.68512189305
0 22.7119559266911
};
\addlegendentry{Without UQ}
\addplot [semithick, darkorange25512714]
table {%
-29 23.4006343738278
-28 23.2172766416839
-27 23.0260762954974
-26 22.8286873414239
-25 22.6531570308732
-24 22.5344570026922
-23 22.4015436118208
-22 22.2140814077384
-21 22.0363838149799
-20 21.8918600603647
-19 21.7436627823318
-18 21.6133886093784
-17 21.5177220847297
-16 21.3980987775278
-15 21.3253298261385
-14 21.2370029484655
-13 21.1699117708724
-12 21.0726443796921
-11 21.0162351953134
-10 20.9793426637453
-9 20.9977937275892
-8 21.0125981186826
-7 21.0430273594775
-6 21.0302074385469
-5 21.0272616810725
-4 21.0029140557422
-3 21.0056771239397
-2 21.0579859995653
-1 21.0941493512567
0 21.0996238957301
};
\addlegendentry{With UQ}
\end{axis}

\end{tikzpicture}}
        \caption{ESD across $t_{lag}$  day for Dataset 1.}
        \label{fig:subfig8}
    \end{subfigure}
     \hspace{1cm}
    \begin{subfigure}{0.35\linewidth} 
        \resizebox{!}{3.5cm}{
\begin{tikzpicture}

\definecolor{darkgray176}{RGB}{176,176,176}
\definecolor{darkorange25512714}{RGB}{255,127,14}
\definecolor{lightgray204}{RGB}{204,204,204}
\definecolor{steelblue31119180}{RGB}{31,119,180}

\begin{axis}[
legend cell align={left},
legend style={fill opacity=0.8, draw opacity=1, text opacity=1, draw=lightgray204},
tick align=outside,
tick pos=left,
x grid style={darkgray176},
y grid style={darkgray176},
xmajorgrids,
ymajorgrids,
xlabel={$t_{lag}$},
xlabel style={font=\large}, 
xmin=-30.45, xmax=1.45,
xtick style={color=black},
y grid style={darkgray176},
ymin=14.4363266925593, ymax=18.6578418691291,
ytick style={color=black},
legend style={font=\large} 
]
\addplot [semithick, steelblue31119180]
table {%
-29 18.4659548156487
-28 18.3783470063015
-27 18.2687163610715
-26 18.1570432010957
-25 18.0487103615564
-24 17.9459238653312
-23 17.8670411593495
-22 17.7345917228012
-21 17.6132384122029
-20 17.4837720973079
-19 17.3067676362321
-18 17.1723937973668
-17 17.0330897134954
-16 16.9145471759065
-15 16.7663500950744
-14 16.6609918435017
-13 16.5607205804582
-12 16.4479105742222
-11 16.3404847356965
-10 16.2246676067192
-9 16.1302954849419
-8 16.0676217669156
-7 16.0036832796006
-6 15.9918595879719
-5 15.9762839330073
-4 15.9633887243603
-3 15.9730956299505
-2 15.9957671083607
-1 16.0190421950471
0 16.0584056043562
};
\addlegendentry{Without UQ}
\addplot [semithick, darkorange25512714]
table {%
-29 17.6802581391853
-28 17.5205269425929
-27 17.4038660078762
-26 17.2674245293429
-25 17.1131427323231
-24 17.0107375059445
-23 16.8190450502884
-22 16.6558717579246
-21 16.5261833739901
-20 16.3292873370271
-19 16.1281621971696
-18 15.9843276661726
-17 15.8152795437953
-16 15.6839249793292
-15 15.4980190972711
-14 15.3427920288419
-13 15.2378479866635
-12 15.1218461466111
-11 14.9972058857955
-10 14.9130524510674
-9 14.8798335984575
-8 14.809710670651
-7 14.7496818214322
-6 14.7123557121627
-5 14.7097740995533
-4 14.6816787973898
-3 14.6717298307975
-2 14.6593688639147
-1 14.6282137460397
0 14.6527762215973
};
\addlegendentry{With UQ}
\end{axis}

\end{tikzpicture}}
        \caption{ESD across $t_{lag}$ for Dataset 2.}
        \label{fig:subfig9}
    \end{subfigure}

    \caption{\footnotesize{ESD for Dataset 1 and Dataset 2 achieved by the two approaches. ESD is evaluated at increasing values of $t_{lag} = t - D$, where $t$ is the day at which the sprouting day is estimated, and $D$ is the actual sprouting day.}}
    \label{fig:stopping_day_comparison}
\end{figure}
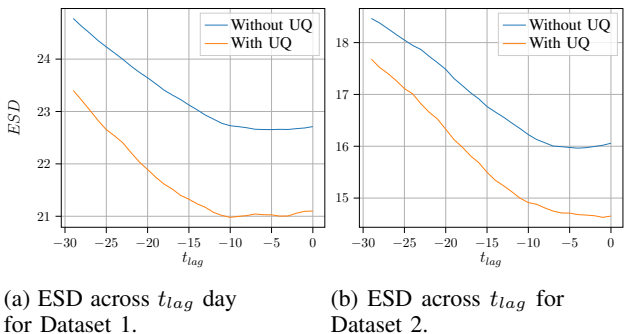

\subsection{Effectiveness of Proposed Approach}
We begin by analyzing the results obtained using our two proposed methodologies, one with uncertainty quantification (UQ) and one without UQ, in terms of mean absolute error (MAE) and error in sprouting day (ESD) across both datasets.
Figures \ref{fig:maesprouting}(a) and \ref{fig:maesprouting}(b) present box plots of the MAE for both approaches across Dataset 1 and Dataset 2. In Dataset 1, the model without UQ yields an average MAE of approximately 45 days, with a maximum of 80 days. Incorporating UQ leads to a modest improvement, reducing the average MAE to 42 days. Additionally, the UQ-enhanced approach demonstrates a lower minimum MAE and a reduced maximum MAE across all plants, indicating a more consistent predictive performance. In Dataset 2, the model without UQ demonstrates an average MAE of 23 days while that leveraging UQ yields an average MAE of 20 days. Consistent with the findings from Dataset 1, the UQ-enhanced approach not only improves the average MAE but also demonstrates better minimum and maximum MAE across all plants. While the improvement remains modest, these results reinforce the benefits of incorporating UQ in enhancing prediction consistency. 
The variation in MAE across datasets may stem from differences in target distributions, variations in sample sizes that capture the signal's variability with differing levels of accuracy, or temperature-dependent physical phenomena reflected in the electrophysiological signals. Further research is needed to distinguish between these possibilities.

We now analyze the ESD for both approaches across the two datasets. As previously discussed, while MAE serves as an indicative measure of prediction quality, it is not entirely sufficient to assess the practical effectiveness of the predictions. This is because, in our use case, decisions are made only once after processing all predictions. Since the model’s output is ultimately used to estimate the sprouting day of each potato tuber, it is crucial to evaluate the accuracy of these predictions directly. 
Figures \ref{fig:maesprouting}(c) and \ref{fig:maesprouting}(d) present a boxplot visualization of the distribution of $ESD_j, j \in [1,N]$ for both datasets, providing a more targeted assessment of prediction reliability. In Dataset 1, incorporating UQ leads to a slightly lower average error in predicted sprouting day (17 days) compared to the model without UQ (19 days). Notably, both approaches achieve a minimum error of 0 days for some plants, indicating that, in certain cases, the predicted sprouting day exactly matches the actual sprouting day. However, both methods also exhibit relatively high maximum errors, reaching up to 65 days, highlighting the presence of challenging cases where predictions significantly deviate from the ground truth. Results from Dataset 2 show similar findings, where the approach leveraging UQ achieves slightly better performance with respect to that without UQ (enhancing average error in sprouting day from 15 to 13), and both approaches are capable of identifying the exact sprouting day for some potato tubers. Despite the wide distribution of errors across all potato tubers, it is noteworthy that both methods exhibit reasonable accuracy for the majority of samples. Specifically, in Dataset 1, most predictions fall within an error margin of 30 days, while in Dataset 2, the majority of errors remain below 20 days. These results indicate that while extreme errors exist, the overall predictive reliability of both approaches remains within an acceptable range for practical use. 

\begin{figure*}[t]
\centering
\setcounter{subfigure}{0} 
\subfloat[]{\includegraphics[width=3.0in]{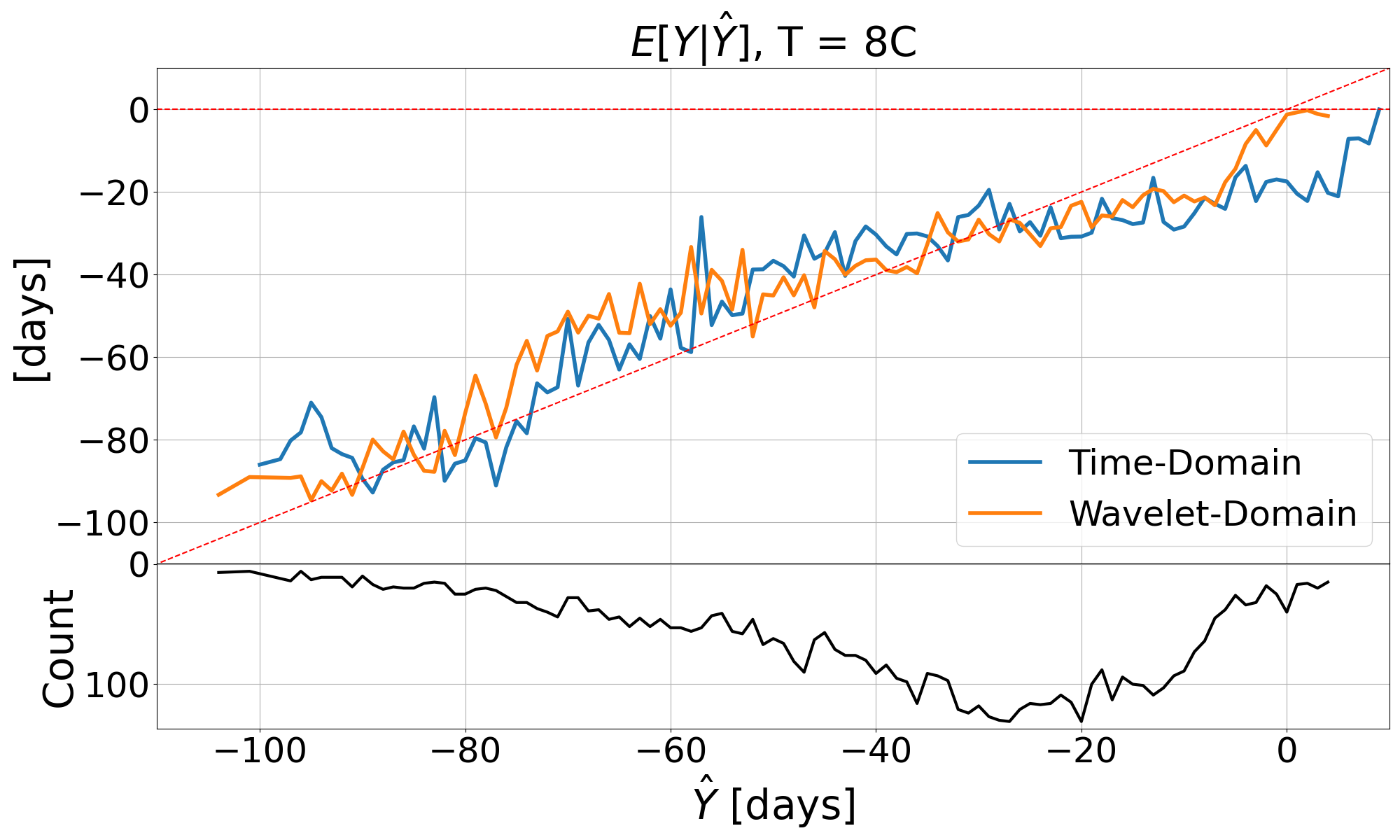}%
\label{fig:8degree-mean}}
\hfil
\subfloat[]{\includegraphics[width=3.0in]{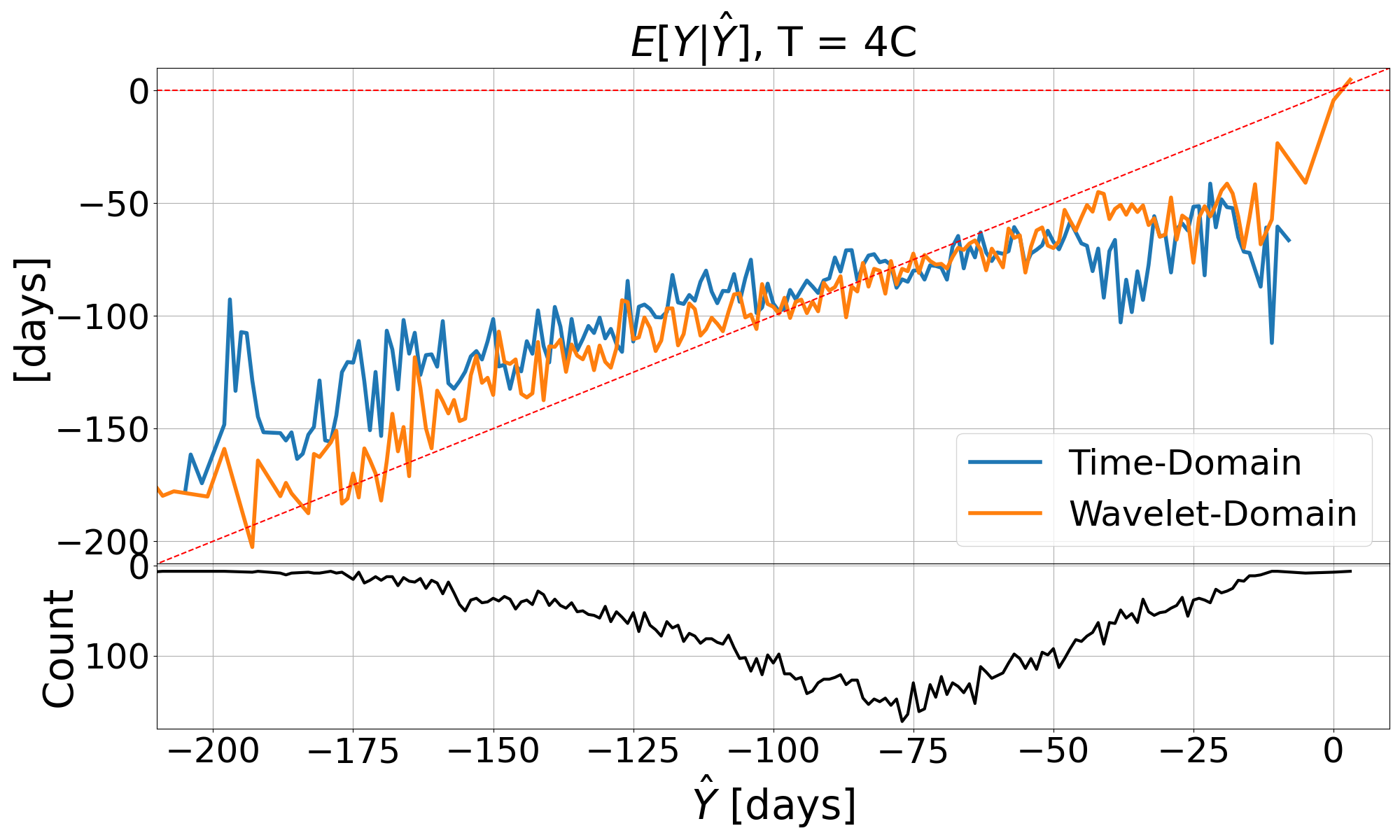}%
\label{fig:8degree-std}}
\hfil
\subfloat[]{\includegraphics[width=3.0in]{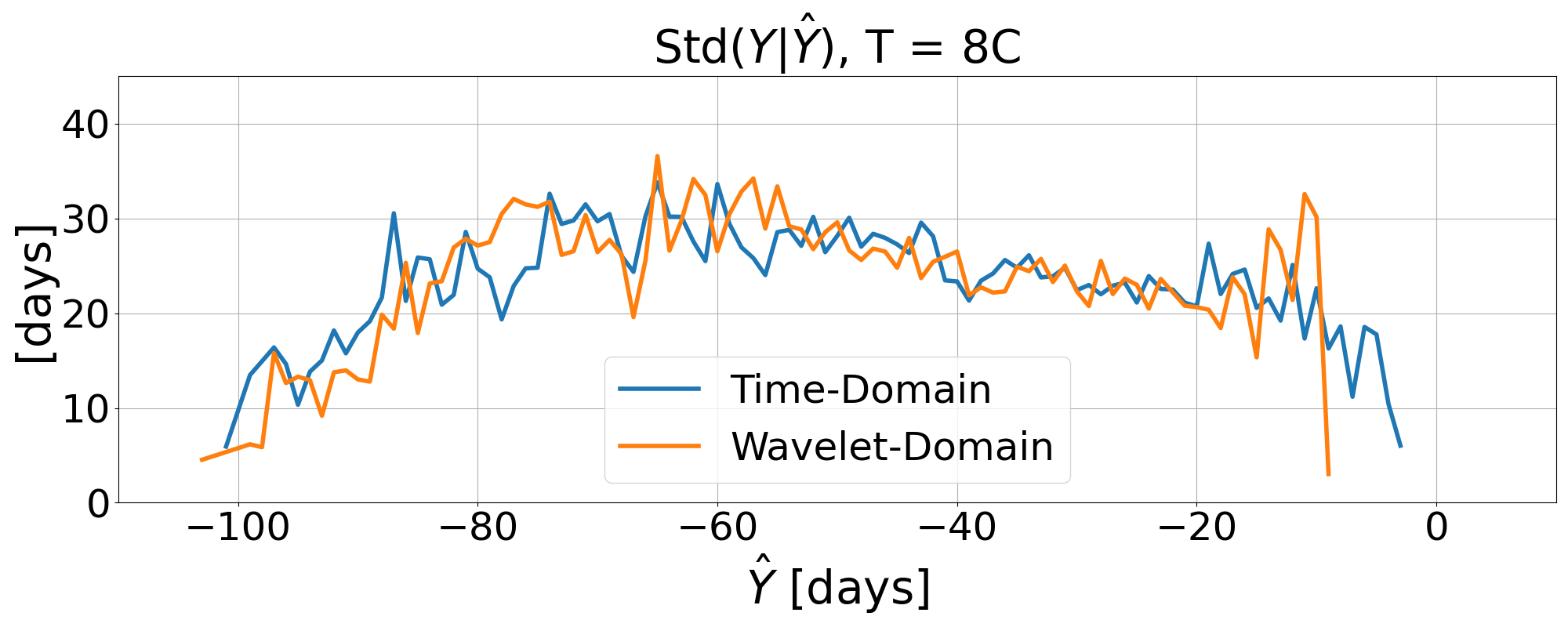}%
\label{fig:4degree-mean}}
\hfil
\subfloat[]{\includegraphics[width=3.0in]{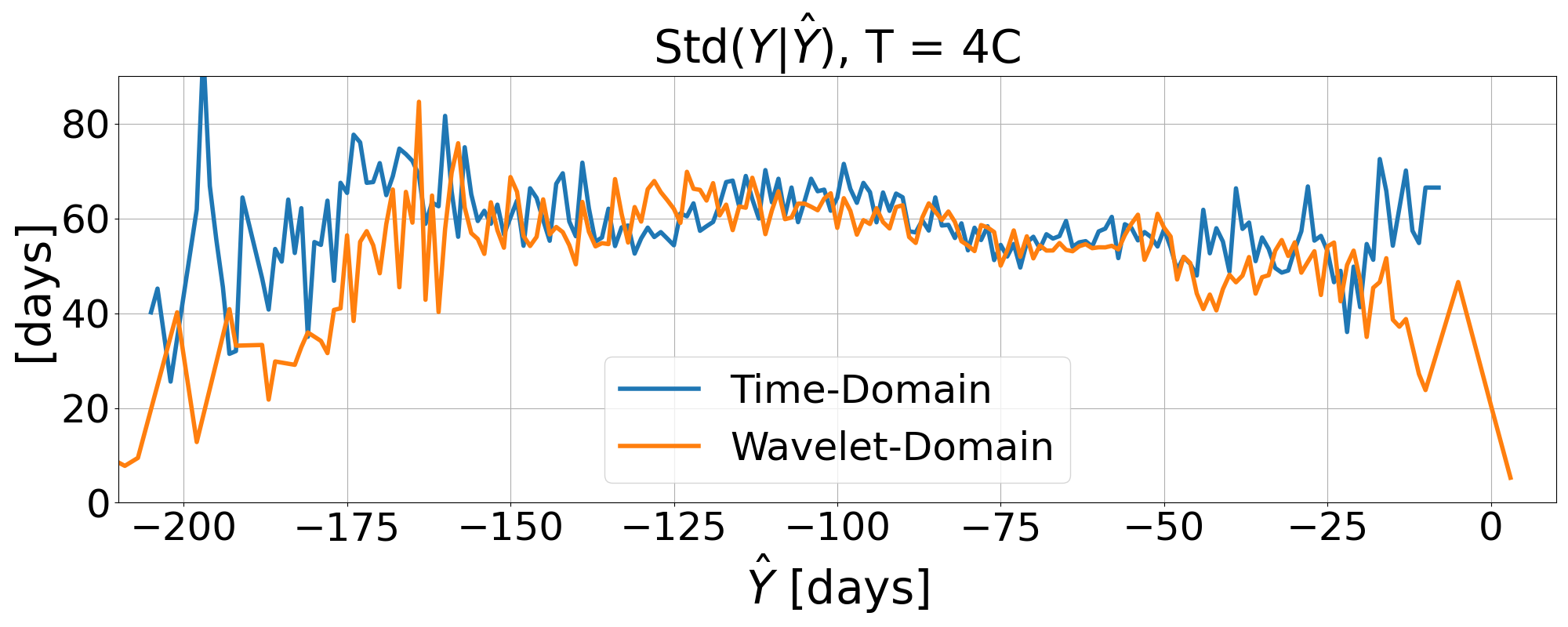}%
\label{fig:4degree-std}}
\caption{\small{(a,b) Calibration plot showing the mean value of the number of days remaining to the sprouting event (indicated as $Y$), conditioned on the value predicted by the model (indicated as $\hat Y$), for potatoes at 8 (dataset 1) and 4 degrees (dataset 2), in the Time and Wavelet domains. In both domains, the same features described in Section \ref{sec:methodology} are employed. In the former, features are extracted from the raw timeseries. In the latter, features are extracted from the Continuous Wavelet Transform applied to the raw time-series. True values align well with the predictions for a wide range of $\hat Y$ values. The distribution of $\hat Y$ is shown on the bottom. (c,d) Similar to before, but instead of the mean, the standard deviation of the exact values conditioned on the predicted values ($Std(Y|\hat Y)$) is reported.}}
\label{fig:results-windows}
\end{figure*}

Alternatively, we can analyze the distribution of $ESD_j, j \in [1,N]$ in terms of the percentile curves across all potato tubers, with and without UQ, in Dataset 1 and Dataset 2 (Figures \ref{fig:percentile_comparison}(a, b)). The approach incorporating UQ consistently demonstrates slightly better performance (i.e., lower error) than the approach without UQ, particularly across the lower percentiles. For instance, in Dataset 1, the approach leveraging UQ show that 40\% of the potato tubers show an error of 16 days or less while the approach without UQ shows that 40\% of tubers are associated with an error of 18 days or less. Similarly, in Dataset 2, the approach leveraging UQ demonstrates that 40\% of potato tubers show an error of 10 days or less, while the approach without UQ show the same percentage of potato tubers associated with an error of 15 days or less. It is important to note that both approaches achieve an error of zero for a subset of potato tubers, which indicates that the model is capable of making perfectly accurate predictions in certain cases. However, the presence of higher errors at the upper percentiles highlights the challenges posed by some difficult-to-predict samples. While incorporating UQ does not eliminate these large deviations, it reduces their magnitude, contributing to improved robustness in the model’s predictions. These findings emphasize the benefits of integrating UQ in reducing prediction errors, particularly in the cases with higher uncertainty. 

Up to now, to validate our model we considered all windows up to the sprouting day, e.g. when estimating the ESD. In a practical scenario, predictions for a potato $j$ must be performed at a time $t < D_j$. Figures \ref{fig:stopping_day_comparison}(a) and (b) present the ESD for each approach in Dataset 1 and Dataset 2, respectively, as a function of a time lag $t_{lag}$. Predictions across potatoes are aligned so that $-t_{lag}$ indicates the number of days up to the sprouting event. Results show a consistent trend across all cases: prediction accuracy improves as the prediction window moves closer to the actual sprouting date, with the error progressively decreasing. However, this improvement stabilizes approximately 10 days before sprouting, where the error reaches a saturation point. This indicates that incorporating additional data beyond this threshold does not significantly enhance prediction accuracy, suggesting that key electrophysiological indicators of sprouting emerge within this critical window. Consequently, this finding can inform practical decision-making by defining an optimal observation period for reliable sprouting predictions while minimizing computational and data collection efforts.

\subsection{Validating Quality of per-window Predictions}
In Fig. \ref{fig:results-windows} we show the quality of individual per-window predictions, under the described LOO-CV framework, by reporting $E[Y|\hat {Y}]$ and $Std(Y|\hat {Y})$ as a function of the predicted $\hat {Y}$ values, for potatoes at $8$ and $4$ degrees separately. Observe the negative sign: a $Y$ value indicates sprouting in $-Y$ days, while $\hat{Y}$ predicts sprouting will occur in $-\hat{Y}$ days. Plots are shown for windows of length 1 day. \\
We first discuss the dependence of $E[Y|\hat{Y}]$ on $\hat{Y}$ values, i.e. the calibration curve, for potatoes at 8 degrees (Fig. \ref{fig:8degree-mean}). This model respects the relation $E[Y|\hat{Y}]=\hat{Y}$ for a wide range of $\hat{Y}$ values, thus leading to well calibrated predictions. Note that a model trained with features with no predictive power would always predict a constant value equal to the mean of the exact targets, i.e. in such a setting $\hat{Y}=E[Y]$. In our case, the range of $\hat{Y}$ values respecting the calibration relation is much larger, proving the suitability of the signals recorded by the sensors to predict the sprouting event. A rolling mean $N_{\text{rolling}}=7$ was used, thus averaging data from a week to compute the final predictions. Without this operation, the calibration curve would have a slightly larger bias with respect to the ideal one. The effect of the rolling mean can be understood in the following way. At times far from the real sprouting date (large $|Y|$) sometimes a fluctuation in the predicted sprouting date could indicate a date much closer to the sprouting event than the exact one. These fluctuations would push down the calibration curve with respect to the ideal situation for small values of $\hat Y$. By performing the rolling mean operation the effect of these fluctuations is minimized and calibration better recovered. The calibration curve for the models at 4 degrees (Fig. \ref{fig:4degree-mean}) showcases a difference between models trained using the wavelet decomposition or without. In particular, a better calibrated model could be obtained when using the wavelet decomposition. 
In this case predictions are well calibrated in the range $-150 \le \hat{Y} \le -50$. Outside from this region ($\hat{Y}<-150$ and $\hat{Y}<-50$) predictions are not calibrated and this should be taken into account when using predictions to take decisions. At variance with models at 8 degrees, this effect could not be mitigated by simply performing a rolling mean over predictions.

We now turn our attention to the plots of the conditional standard deviations (Fig. \ref{fig:8degree-std} and \ref{fig:4degree-std}). The maximal standard deviation is relatively high, around $30$ and $60$ days for potatoes at $8$ and $4$ degrees respectively. Therefore an individual prediction on a single potato can identify the sprouting event but with a high uncertainty. This suggests to combine signals of different potatoes and at different times to improve accuracy in a decision setting. Interestingly, the standard deviation decreases from $30$ to $20$ days for potatoes at 8 degrees and from $60$ to $40$ days for potatoes at $4$ degrees as $\hat{Y}$ values approach sprouting. In other words, the closer the model predicts sprouting, the more the model predictions can be trusted. Finally, the standard deviation is lower near and far from sprouting, making these regions more informative. 

\section{Conclusion}\label{conclusion}
Our work explores the use of electrophysiological sensor data with machine learning techniques to predict potato sprouting before any visible signs appear. To this end, we conduct experiments for recording electrophysiological data from potato tubers and use this data to develop an approach encompassing both signal processing and predictive modeling for predicting potato sprouting. Experimental results from two datasets, covering potatoes stored under different conditions, demonstrate that combining plant electrophysiology with machine learning offers a promising approach for early potato sprouting detection. 
On a scientific level, interpretability methods can help us understand the signal patterns linked to sprouting, allowing us to refine our models by prioritizing signal windows with higher information content. This would enhance our initial attempt to weight signal predictions based on uncertainty, leading to incremental improvements. From an industrial perspective, developing a model that optimally aggregates predictions from multiple potatoes and provides insights into the distribution of sprouting days within a batch of potatoes will enhance the practical applicability of our contribution in an industrial setting.


\ifCLASSOPTIONcaptionsoff
  \newpage
\fi

\end{document}